\pdfoutput=1

\documentclass{article} % For LaTeX2e
\usepackage{iclr2020_conference,times}

\usepackage{iclr2020_conference_arxiv,times}
\usepackage{amsmath,amsfonts,bm}

\def\eqref#1{equation~\ref{#1}}
\def\1{\bm{1}}

\DeclareMathAlphabet{\mathsfit}{\encodingdefault}{\sfdefault}{m}{sl}
\SetMathAlphabet{\mathsfit}{bold}{\encodingdefault}{\sfdefault}{bx}{n}

\DeclareMathOperator*{\argmin}{arg\,min}

\usepackage{hyperref}
\usepackage{url}

\usepackage{natbib}

\usepackage[T1]{fontenc}    % use 8-bit T1 fonts
\usepackage[utf8]{inputenc} % allow utf-8 input
\usepackage{hypernat}
\usepackage{booktabs} 
\usepackage{amsfonts}       % blackboard math symbols
\usepackage{nicefrac}  
\usepackage{microtype} 
\usepackage{tikz}
\usepackage{float}
\usepackage{graphicx}
\usepackage{subcaption}
\usepackage{stackengine}
\usepackage{amsmath} % For typesetting math
\usetikzlibrary{positioning}
\usetikzlibrary{shapes.geometric, arrows}
\usepackage{multirow}
\usepackage{enumitem}
\usepackage{algorithm}
\usepackage{algorithmic}
\usepackage[justification=centering]{caption}
\usepackage{todonotes}
\usepackage{mathtools}

\usepackage{siunitx}

\usepackage{tikz}
\usetikzlibrary{decorations.pathreplacing,calc}
\newcommand{\tikzmark}[1]{\tikz[overlay,remember picture] \node (#1) {};}

\newcommand*{\AddNote}[4]{%
    \begin{tikzpicture}[overlay, remember picture]
        \draw [decorate,decoration={brace,amplitude=0.5em},xshift=15.0cm,thick,black]
            ($(#3)!(#1.north)!($(#3)-(0,1)$)$) --  
            ($(#3)!(#2.south)!($(#3)-(0,1)$)$)
                node [align=center, text width=3cm, pos=0.5, anchor=west] {#4};
    \end{tikzpicture}
}%

\usepackage{amssymb} % Adds new symbols to be used in math
\usepackage{amsthm}
\usepackage{epstopdf} 
\usepackage{float}

\newcommand{\cA}{\mathcal{A}}

\newcommand{\cD}{\mathcal{D}}

\newcommand{\cG}{\mathcal{G}}

\newcommand{\cL}{\mathcal{L}}

\newcommand{\cP}{\mathcal{P}}

\newcommand{\cS}{\mathcal{S}}

\newcommand{\Real}{\mathbb{R}}

\newcommand{\Esp}{\mathbb{E}}

\newcommand{\old}{\text{old}}
\newcommand{\kl}{\text{D}_{\text{KL}}}
\newcommand{\ar}{\text{AR}}

\newcommand{\rollout}{\texttt{rollout}}
\newcommand{\updatearchive}{\texttt{update\_archive}}
\newcommand{\withnoise}{\text{with noise}}
\newcommand{\withoutnoise}{\text{without noise}}

\newcommand{\episode}[1]{\text{over}~#1~\text{episode}}
\newcommand{\episodes}[1]{\text{over}~#1~\text{episodes}}

\renewcommand{\vec}[1]{\ensuremath{\mathbf{#1}}}

\newcommand{\mat}[1]{\ensuremath{\mathbf{#1}}}

\usepackage{xspace}
\newcommand{\method}{ARAC\xspace}

\newcommand{\SN}[1]{\num[{round-mode = places,round-precision     = 2,scientific-notation = fixed,fixed-exponent=0, separate-uncertainty = true}]{#1}}
\title{
Attraction-Repulsion Actor-Critic for Continuous Control Reinforcement Learning}

\author{
 \large \noindent Thang Doan\thanks{These authors contributed equally.}\\
  McGill University, Mila\\
  \And
  \large Bogdan Mazoure\textsuperscript{\text{*} }\\
  McGill University, Mila   \\
  \And
  \large Moloud Abdar\\
  Deakin University\\
  \And
  \large Audrey Durand \\
  Université Laval, Mila\\
  \And
    \large Joelle Pineau\\
  McGill University, Mila\\ Facebook AI Research \\
  \And 
  \large R Devon Hjelm\\
  MSR Montreal, Mila \\Université de Montréal \\
}

\begin{document}
\maketitle
% \footnotetext[1]{Corresponding author: thang.doan@mail.mcgill.ca, Desautels Faculty of Management, McGill University}
% \footnotetext[2]{INRS-EMT, Universit\'e du Qu\'ebec} \footnotetext[3]{Department of Mathematics \& Statistics, McGill University} \footnotetext[4]{School of Computer Science, McGill University} \footnotetext[5]{Facebook AI Research} \footnotetext[6]{Microsoft Research}\footnotetext[7]{MILA}
 \begin{abstract}
Continuous control tasks in reinforcement learning are important because they provide an important framework for learning in high-dimensional state spaces with deceptive rewards, where the agent can easily become trapped into suboptimal solutions.
One way to avoid local optima is to use a population of agents to ensure coverage of the policy space, yet learning a population with the ``best" coverage is still an open problem. In this work, we present a novel approach to population-based RL in continuous control that leverages properties of normalizing flows to perform attractive and repulsive operations between current members of the population and previously observed policies. Empirical results on the MuJoCo suite demonstrate a high performance gain for our algorithm compared to prior work, including Soft-Actor Critic (SAC). 

\end{abstract}

\section{Introduction}
Many reinforcement learning (RL) tasks such as robots and self-driving cars pose a major challenge due to large action and state spaces~\citep{lee2018deep}. In particular, environments with large non-convex continuous action spaces are prone to \emph{deceptive} rewards, i.e. local optima \citep{novelty_seeking_agents}. Applying traditional policy optimization algorithms to these domains often leads to locally optimal, yet globally sub-optimal policies. This implies that learning should involve some form of exploration.

Not all RL domains that require exploration are suitable for understanding how to train agents that are robust to deceptive rewards.
For example, Montezuma Revenge, a game in the Atari Learning Environment~\citep{ale}, has \emph{sparse rewards}; algorithms that perform the best on this task encourage exploration by providing learning signal to the agent~\citep{count_based_exploration}.

% in this setting do so as a means of providing more consistent learning signal for the agent~\citep{count_based_exploration}. 
On the other hand, continuous control problems, such as MuJoCo~\citep{mujoco}, already provide the agent with a dense reward signal. Yet, the high-dimensional action space induces a multimodal (potentially deceptive) reward landscape. Such domain properties can lead the agent to sub-optimal policies. 
For example, in the biped environments, coordinating both arms and legs is crucial for performing well on even simple tasks such as forward motion. However, simply learning to maximize the reward can be detrimental in the long run: agents will tend to run and fall further away from the start point instead of discovering a stable walking motion. Exploration in this setting serves to provide a more reliable learning signal for the agent by covering more different types of actions during learning.

One way to maximize action space coverage is the maximum entropy RL framework \citep{ziebart2010modeling}, which prevents variance collapse by adding a policy entropy auxiliary objective. One such prominent algorithm, Soft Actor-Critic (SAC,\cite{sac}), has been shown to excel in large continuous action spaces. To further improve on exploration properties of SAC, one can maintain a population of agents that cover non-identical sections of the policy space. To prevent premature convergence, a diversity-preserving mechanism is typically put in place; balancing the objective and the diversity term becomes key to converging to a global optimum~\citep{diversity_driven_exploration}. This paper studies a particular family of population-based exploration methods, which conduct coordinated local search in the policy space. Prior work on population-based strategies improves performance on continuous control domains through stochastic perturbation on a single actor's parameter~\citep{cem_rl} or a set of actor's parameters~\citep{novelty_seeking_agents, stein_variationam_pg,erl}. We hypothesize that exploring directly in the policy space is more important than perturbing the parameters of the policy, as the latter does not guarantee diversity (i.e., different neural network parameterizations can approximately represent the same function).

Given a population of RL agents, we enforce local exploration using an Attraction-Repulsion (AR) mechanism. The later consists in adding an auxiliary loss to encourage pairwise attraction or repulsion between members of a population, as measured by a divergence term. We make use of the Kullback-Leibler (KL) divergence because of its desirable statistical properties and its easiness of computation. However, naively maximizing the KL term between two Gaussian policies can be detrimental (e.g. drives both means apart). As a result, we parametrize the policy with a general family of distributions called Normalizing Flows (NF)~\citep[NFs,][]{rezende2015variational}; this modification allows to improve upon AR+Gaussian (see Appendix Figure~\ref{fig:evo_gaussian}). NFs are shown to improve the expressivity of the policies using invertible mappings while maintaining entropy guarantees \citep{boosting_trpo, sac_nf}. Nonlinear density estimators have been previously used for deep RL problems in contexts of distributional RL~\citep{gan_dqn} and reward shaping with count-based methods~\citep{count_based_exploration}. The AR objective blends particularly well with SAC, since computing the KL requires stochastic policies with tractable densities for each agent.

\section{Preliminaries}
\label{sec:prelim}

We first formalize the RL setting in a Markov decision process (MDP).
A discrete-time, finite-horizon, MDP~\citep{bellman1957markovian,puterman2014markov}
is described by a state space $\cS$, an action space $\cA$, a transition function $\cP:\cS\times \cA \times \cS\mapsto \Real^+$, and a reward function $r:\cS\times \cA\mapsto \Real$.\footnote{$\cA$ and $\cS$ can be either discrete or continuous.}
On each round $t$, an agent interacting with this MDP observes the current state $s_t\in\cS$, selects an action $a_t\in\cA$, and observes a reward $r(s_t, a_t) \in\Real$ upon transitioning to a new state $s_{t+1}\sim \cP(s_t, a_t)$.
Let $\gamma\in[0,1]$ be a discount factor.
The goal of an agent evolving in a discounted MDP is to learn a policy $\pi:\cS\times\cA\mapsto[0,1]$ such as taking action $a_t\sim\pi(\cdot|s_t)$ would maximize the expected sum of discounted returns, 
\begin{align*}
V^\pi(s) = \Esp_\pi \bigg[\sum_{t=0}^\infty \gamma^t r(s_t, a_t) | s_0 = s\bigg].
\end{align*}
% The corresponding state-action value function can be written as the expected discounted rewards from taking action $a$ in state $s$, that is
% \begin{align}
% Q_t^\pi(s,a)=\Esp_{\pi}\bigg[\sum_{i=t}^\infty\gamma^{i-t}r(s_i,a_i)|s_t=s,a_t=a\bigg].
% \end{align}
In the following we use $\rho_\pi$ to denote the trajectory distribution induced by following policy $\pi$.
If $\cS$ or $\cA$ are vector spaces, action and space vectors are respectively denoted by $\vec{a}$ and $\vec{s}$.

\subsection{Discovering new solutions through population-based Attraction-Repulsion}

Consider evolving a population of $M$ agents, also called \textit{individuals}, $\{ \pi_{\theta_m} \}_{m=1}^{M}$, each agent corresponding to a policy with its own parameters.
In order to discover new solutions, we aim to generate agents that can mimic some target policy while following a path different from those of other policies.

Let $\cG$ denote an archive of policies encountered in previous generations of the population.
A natural way of enforcing $\pi$ to be different from or similar to the policies contained in $\cG$ is by augmenting the loss of the agent
% of the actor (Eq.~\ref{eq:sac_pi_loss})
with an Attraction-Repulsion (AR) term:
\begin{align}
\cL_{\ar}= -\underset{{\pi' \sim \cG}}{\Esp}\big[\beta_{\pi'}\kl[\pi || \pi']\big],
\label{eq:attraction_repulsion}
\end{align}
where $\pi'$ is an archived policy 
% sampled uniformly from the archive $\cG$
and $\beta_{\pi'}$ is a coefficient weighting the relative importance of the Kullback-Leibler (KL) divergence between $\pi$ and $\pi'$ which we will choose to be a function of the average reward (see Sec. \ref{sec:ar_function} below).
Intuitively, Eq.~\ref{eq:attraction_repulsion} adds to the agent objective the average distance between the current and the archived policies.
For $\beta_{\pi'} \geq 0$, the agent tends to move away from the archived policy's behavior (i.e. \emph{repulsion}, see Figure~\ref{fig:illustration_alg}) a). On the other hand, $\beta_{\pi'} < 0$ encourages the agent $\pi$ to imitate $\pi'$ (i.e. \emph{attraction}).

\paragraph{Requirements for AR}

In order for agents within a population to be trained using the proposed AR-based loss (Eq.~\ref{eq:attraction_repulsion}), we have the following requirements:
\begin{enumerate}
% \item Agents should be described by parametric policies, so that individual policies [working]
\item Their policies should be stochastic, so that the KL-divergence between two policies is well-defined.
\item Their policies should have tractable distributions, so that the KL-divergence can be computed easily, either with closed-form solution or Monte Carlo estimation.
\end{enumerate}
Several RL algorithms enjoy such properties~\citep{ppo, trpo, sac}. 
In particular, the soft actor-critic~\citep[SAC,][]{sac} is a straightforward choice, as it currently outperforms other candidates and is off-policy, thus maintains a single critic shared among all agents (instead of one critic per agent), which reduces computation costs.

\subsection{Soft actor-critic}

SAC~\citep{sac} is an off-policy learning algorithm which finds the information projection of the Boltzmann Q-function onto the set of diagonal Gaussian policies $\Pi$:
\begin{align*}
    \pi=\argmin_{\pi'\in \Pi}  \kl\bigg(\pi'(.|\vec{s}_t) \bigg\lVert \frac{\exp{(\frac{1}{\alpha}Q^{\pi_\old}(\vec{s}_t,.))}}{Z^{\pi_\old}(\vec{s}_t)} \bigg),
    %\label{eq:sac_pi_loss}
\end{align*}
where $\alpha\in (0,1)$ controls the temperature, i.e. the peakedness of the distribution. 
% When $\mat{\Sigma}$ is diagonal, then the entropy is directly proportional to the sum of log-variances. 
% The task tackled by SAC can hence be re-formulated as maximizing expected discounted rewards while keeping the volume of the policy at a certain level. Doing so keeps exploration active and prevents mode collapse on the highest reward.
The policy $\pi$, critic $Q$, and value function $V$ are optimized according to the following loss functions:
\begin{align}
    \label{eq:sac_pi_loss}
    \mathcal{L}_{\pi, \text{SAC}} &= \Esp_{\vec{s}_t\sim \mathcal{B}}[\Esp_{\vec{a}_t\sim\pi}[\alpha \log \pi(\vec{a}_t|\vec{s}_t)-Q(\vec{s}_t,\vec{a}_t)]] \\
% \mathcal{L}_{Q} &= \underset{(s,a,r,s') \sim \mathcal{B}  }{\Esp}\big[(Q(s,a)-(r+\gamma Q(s',\pi(.|s'))))^2\big]
\mathcal{L}_{Q} &= \underset{(s,a,r,s') \sim \mathcal{B}  }{\Esp}\big[\{Q(s,a)-(r+\gamma V_\nu^\pi(s') )\}^2\big]
\label{eq:critic_loss}\\
\begin{split}
    \cL_V &= \Esp_{\vec{s}_t \sim \cD}\bigg[\frac{1}{2}\big\{V_\nu^\pi(\vec{s}_t)  -\Esp_{\vec{a}_t\sim \pi}[Q(\vec{s}_t,\vec{a}_t)-\alpha\log\pi(\vec{a}_t|\vec{s}_t)]\big\}^2\bigg]
\end{split}
     \label{eq:v_loss},
\end{align}
where $\mathcal{B}$ is the replay buffer.
The policy used in SAC as introduced in \citet{sac} is Gaussian, which is both stochastic and tractable, thus compatible with our AR loss function in Eq.~\ref{eq:attraction_repulsion}.
Together with the AR loss in Eq.~\ref{eq:attraction_repulsion}, the final policy loss becomes:
\begin{align}
    \label{eq:sac_pi_ar_loss}
    \cL_{\pi}=\cL_{\pi, \text{SAC}} + \cL_{\text{AR}} 
\end{align}

However, Gaussian policies are arguably of limited expressibility; we can improve on the family of policy distributions without sacrificing qualities necessary for AR or SAC by using Normalizing Flows~\citep[NFs,][]{rezende2015variational}.

\subsection{Normalizing flows}
\label{sec:prelim:nfs}

NFs~\citep{rezende2015variational} were introduced as a means of transforming simple distributions into more complex distributions using learnable and invertible functions.
Given a random variable $\vec{z}_0$ with density $q_0$, they define a set of differentiable and invertible functions, $\{f_i\}_{i=1}^{N}$, which generate a sequence of $d$-dimensional random variables, $\{\vec{z}_i\}_{i=1}^N$.
% \begin{align}
%     \log q_j(\vec{z}_j)= \log q_0(\vec{z}_0)- \sum_{i=1}^j\log \Bigl| \text{det} \frac{\partial f_i(\vec{z}_{i-1})}{\partial \vec{z}_{i-1}} \Bigr|.
% \end{align}

%
% Specific forms of the mapping can be selected based on desired properties: highly expressive and sometimes invertible maps~\citep{rezende2015variational}, always invertible affine transformations~\citep{iaf}, volume-preserving and orthogonal transformations~\citep{tomczak2016improving} or always invertible and highly expressive networks~\citep{naf}.
% Due to the high time complexity of RL algorithms, it is preferable to maximally reduce the dimensionality of the actor models while keeping the sample complexity on-par. For this reason, we propose to use the highly expressive and yet very light family of radial contractions around a point $\vec{z}_0$, originally introduced by \cite{rezende2015variational} and defined as
%
Because SAC uses explicit, yet simple parametric policies, NFs can be used to transform the SAC policy into a richer one (e.g., multimodal) without risk loss of information.
For example, \citet{sac_nf} enhanced SAC using a family of radial contractions around a point $\vec{z}_0\in\Real^d$,
\begin{align}
    f(\vec{z})=\vec{z}+\frac{\beta}{\alpha+||\vec{z}-\vec{z}_0||_2}(\vec{z}-\vec{z}_0)
\end{align}
% with a closed-form determinant
% \begin{align}
%     \Bigl| \det\frac{\partial f(\vec{z})}{\partial\vec{z}}\Bigr|
%     &= [1+\beta h(\alpha,r)]^{d-1} \nonumber \\
%     &\qquad \cdot [1+\beta h(\alpha,r)+\beta h'(\alpha,r)r]
% \end{align}
for $\alpha\in \mathbb{R}^{+}$ and $\beta\in \mathbb{R}$.
% Inspired by [CITE PREPRINT], we make use of highly expressive and light (parameter-wise) radial policies learned through soft actor-critic updates.
This results in a rich set of policies comprised of an initial noise sample $\vec{a}_0$, a state-noise embedding $h_\theta(\vec{a}_0,\vec{s}_t)$, and a flow $\{f_{\phi_i}\}_{i=1}^N$ of arbitrary length $N$, parameterized by $\phi=\{\vec{\phi}_i\}_{i=1}^N$. Sampling from the policy $\pi_{\phi,\theta}(\vec{a}_t|\vec{s}_t)$ can be described by the following set of equations:
\begin{align}
\label{eq:sac_nf}
\begin{split}
    \vec{a}_0 &\sim \mathcal{N}(0,\mat{I});\\
    \vec{z} &= h_\theta(\vec{a}_0,\vec{s}_t);\\
    \vec{a}_t &= f_{\phi_{N}} \circ f_{\phi_{N-1}} \circ...  \circ f_{\phi_{1}} (\vec{z}),
\end{split}
\end{align}
where $h_{\theta}=\vec{a}_0 \vec{\sigma}\mat{I} + \mu(\vec{s}_t)$ depends on the state and the noise variance $\sigma>0$.
%, for which the later can be taken conditional $\sigma(\vec{s})$, or average $\sigma$.
Different SAC policies can thus be crafted by parameterizing their NFs layers.
% \rdh{I don't understand this statement}.
% \rdh{Also I don't think we should go into normalizing flows in too much detail. It doesn't seem to be that important other than it's a way to make the policy a little more rich. We just rely on some paper (ours that got rejected) that noone has heard about. There aren't even experiments that show that NF helps!}
% \audrey{It may be a good idea to state explictely what is optimized in SAC-NF (can refer to Eq.3, 4, 5 if relevant). We cannot expect people to go read the SAC-NF preprint in order to understand this work.}\bogdan{corrected}

% \rdh{Why isn't evolutionary algorithms AR background? Eqs 9, 10 should probably be here!}

\section{\method : Attraction-Repulsion Actor-Critic}
\label{sec:methodology}

\begin{figure}[t]
    \centering
    \includegraphics[width=.65\linewidth]{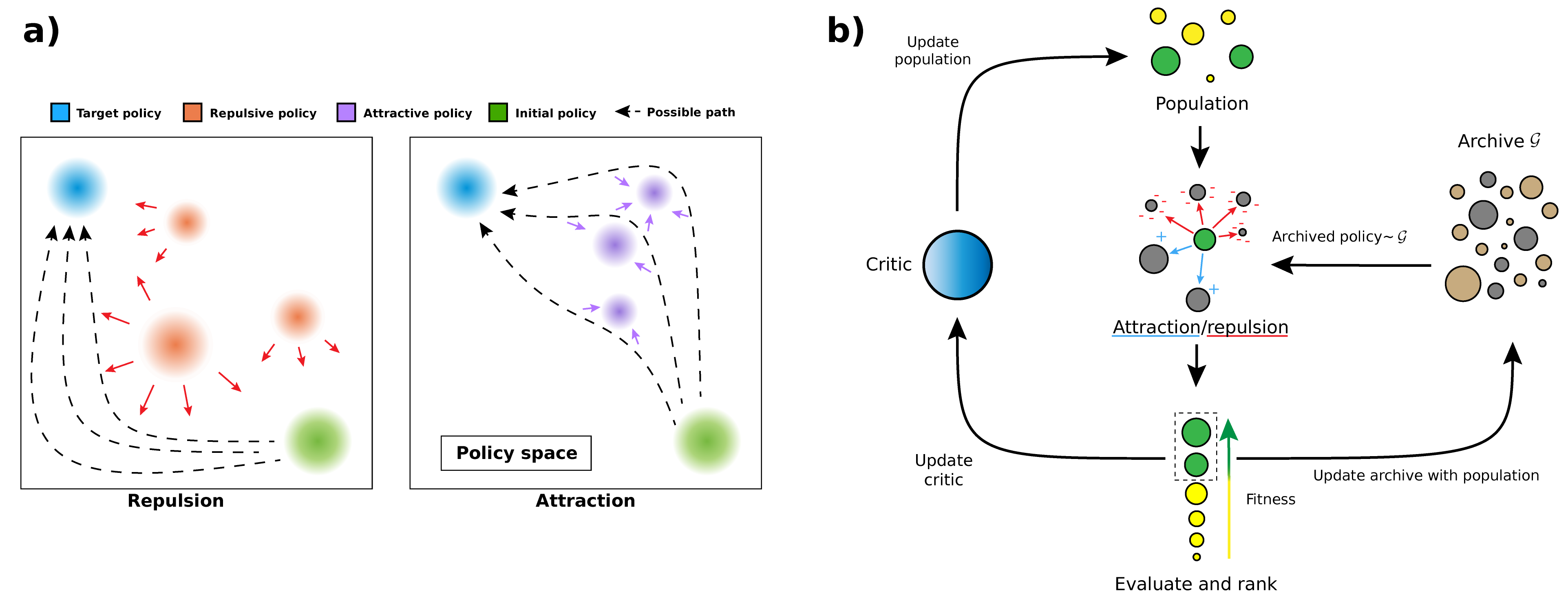}
    \caption{\textbf{a)} Augmenting the loss function with AR constraints allows an agent to reach a target policy by following different paths. Attractive and Repulsive policies represent any other agent's policy. \textbf{b)} General flow of the proposed \method strategy.
    }
    \label{fig:illustration_alg}
\end{figure}

% We now introduce the \method algorithm, which uses
% % combines NF to derive an efficient exploration strategy. More specifically, \method extends the existing
% SAC-NF~\citep{sac_nf} agents within an evolutionary-flavoured strategy to improve the coverage of the policy space.
We now detail the general procedure for training a population of agents using the proposed diversity-seeking AR mechanism. More specifically, we consider here SAC agents enhanced with NFs~\citep{sac_nf}.
Figure~\ref{fig:illustration_alg} displays the general flow of the procedure. Algorithm~\ref{alg:evo_nf} (Appendix) provides the pseudo-code of the proposed \method strategy, where sub-procedures for rollout and archive update can be found in the Appendix.

\paragraph{Overview}

\method works by evolving a population of $M$ SAC agents $\{ \pi_{\phi,\theta}^{m} \}_{m=1}^{M}$
% \rdh{we were overriding N, i, etc, so I changed this so it's less confusing, but I probably missed some places}
with radial NFs policies
% \rdh{do they share parameters?}\bogdan{they are completely independent parameter-wise, but they share 1 critic}
(Eq.~\ref{eq:sac_nf}) and shared critic $Q_{\omega}$, and by maintaining an archive of policies encountered in previous generations of the population.
% \footnote{We found a shared critic works effectively while being memory-efficient.} \rdh{see footnote}
% Each actor in the population, also called an \textit{individual}, corresponds to a unique policy.
After performing $T$ steps per agent on the environment~(Alg.~\ref{alg:evo_nf} L\ref{alg:ln:beg_samples}-\ref{alg:ln:end_samples}), individuals are evaluated by performing $R$ rollouts\footnote{These steps can be performed in parallel.} on the environment~(Alg.~\ref{alg:evo_nf} L\ref{alg:ln:beg_eval}-\ref{alg:ln:end_eval}).
% \rdh{correct?} \thang{yes, after every 10,000 steps, we perform 10 rollouts with each individuals to rank them}
% after which they are ranked based on their \textit{fitness}~(Alg.~\ref{alg:evo_nf} L23-25).
This allows to identify the top-$K$ best agents~ (Alg.~\ref{alg:evo_nf} L\ref{alg:ln:topk}), also called \textit{elites}, which will be used to update the critic as they provide the most meaningful feedback~(Alg.~\ref{alg:evo_nf} L\ref{alg:ln:beg_update_critic}-\ref{alg:ln:end_update_critic}).
% \rdh{I don't follow why it needs to be in-order} \thang{what do you mean by "in order"?, the intuition is basically update the critics with the best agents: they provide the next action pi(s') in equation (4)  }\bogdan{in-order is meant as in "to ensure that", not in that specific order}\audrey{Changed it to avoid confusion}.
The archive is finally updated in a diversity-seeking fashion using the current population~(Alg.~\ref{alg:evo_nf} L\ref{alg:ln:archive}).

The core component of the proposed approach lies within the update of the agents~(Alg.~\ref{alg:evo_nf} L\ref{alg:ln:beg_update_actors}-\ref{alg:ln:end_update_actors}). During this phase, elite individuals are updated using AR operations w.r.t. policies sampled from the archive (Eq.~\ref{eq:sac_pi_ar_loss}), whereas non-elites are updated regularly (Eq.~\ref{eq:sac_pi_loss}).
% \audrey{Validate Alg. line numbers and Eq. numbers cited previously.} \thang{done}
% Genetic operators defined in such a way allow to increase the overall diversity of the population and lead to discovery of new solutions. 

% \subsection{Picking the elites for updating the critic}

% As in classic evolutionary strategies \citep{es_book_1989}, elite individuals are used to obtain the next generation of individuals. Every $X$ environment interaction steps, $R$ rollouts are performed to evaluate and rank each individual by their fitness, here defined as the cumulative reward $\sum_{t} r(s_t,a_t)$ observed over the $R$ rollouts. The pseudo-code of the rollout procedure is provided in Appendix (Alg.~\ref{alg:rollout}). \thang{Is is better the idea of elites?} The top-$K$ most fit individuals of the current generation are used to update the critic $Q_\omega$:
% %
% \begin{align}
% \label{eq:critic_with_elites}
% \cL_{Q}= \underset{(s,a,r,s') \sim \cB  }{\Esp}\big[(Q(s,a)-(r+\gamma Q(s',\pi(.|s'))))^2\big], \pi \in \cE,
% \end{align}
% where $\cB$ is the buffer that stores transitions $(s,a,r,s')$ and $\pi$ is an elite policy.

\subsection{Enhancing diversity in the archive}

% \thang{we should tell that reward is not equivalent of diversity, but it would be to time consuming to compute a KL divergence between policy as a diversity proxy, so instead we try to have policy of different reward}
% \audrey{Are you saying that the KL is actually never computed? In this case, do we still require an RL algorithm with tractable distributions?}\bogdan{The ranking is based on average reward over N episodes, but really it should be the KL of the policy from the optimal policy. We just don't know the optimal one, so this reward is a proxy}
Throughout the training process, we maintain an archive $\cG$ of maximum capacity $G$, which contains some previously encountered policies. The process goes as follow: until reaching full capacity, the archive saves a copy of the parameters of every individual in the population after the evaluation step. However, by naively adding all individuals as if the archive were just a heap, the archive could end up filled with policies leading to similar rewards, which would result in a loss of diversity~\citep{mauldin1984maintaining}. We mitigate this issue by keeping track of two fitness clusters (low and high) using the partition formed by running a $k$-means algorithm on the fitness value. Hence, when $|\cG|=G$ is reached and a new individual is added to the archive, it randomly replaces an archived policy from its respective cluster. This approach, also known as \textit{niching}, has proved itself effective at maintaining high diversity levels~\citep{mahfoud1995niching,gupta2012overview}.
% In our case, maintaining diversity in the archive is important as these archived policies will serve as anchors for crafting new diverse policies.

\subsection{Discovering new policies through Attraction-Repulsion}
\label{sec:ar_function}
The crux of this work lies in the explicit search for diversity in the policy space achieved using the AR mechanism. Since the KL between two base policies (i.e. input of the first flow layer) can be trivially maximized by driving their means apart, we apply attraction-repulsion only on the flow layers, while holding the mean of the base policy constant. This ensures that the KL term doesn't depend on the difference in means and hence controls the magnitude of the AR mechanism.
Every time the AR operator is applied (Alg.~\ref{alg:evo_nf} L\ref{alg:ln:beg_ar}-\ref{alg:ln:end_ar}), $n$ policies are sampled from the archive and are used for estimating the AR loss (Eq.~\ref{eq:attraction_repulsion}). As in \cite{diversity_driven_exploration}, we consider two possible strategies to dictate the value of $\beta_{\pi'}$ coefficients for policies $\pi'\sim \cG$:
% Each time archived policies are sampled to apply the AR operator (Alg.~\ref{alg:evo_nf} L17-18), the $\beta_{\pi^{a}}$ coefficients of each sampled policy $\pi^a$ are updated. As in \cite{diversity_driven_exploration}, we consider two possible strategies (parameterizing the algorithm) to dictate the update of coefficients:
\begin{align}
\label{eq:proactive}
\beta_{\pi'} &= -\bigg[2\bigg(\frac{f(\pi')-f_{min}}{f_{max}-f_{min}}-1\bigg)\bigg] & \quad \text{(proactive)} \\
\label{eq:reactive}
\beta_{\pi'} &= 1-\frac{f(\pi')-f_{min}}{f_{max}-f_{min}} & \quad \text{(reactive)} 
\end{align}
where $f(\pi)\footnote{We overload our notation f for both the normalizing flow and the fitness depending on the context}$
% \rdh{we're overriding $f$ from normalizing flows!} \bogdan{I changed the "f" from NF to "q" because of the link with VI}
represents the fitness function of policy $\pi$ (average reward in our case), and $f_{min}$ and $f_{max}$ are estimated based on the $n$ sampled archived policies. The \textit{proactive} strategy aims to mimic high reward archived policies, while
% moving away from low reward regions. On the other hand,
the \textit{reactive} strategy is more cautious, only repulsing away the current policy from low fitness archived policies. Using this approach, the current agent policy will be attracted to some sampled policies ($\beta_{\pi'} < 0$) and will be repulsed from others ($\beta_{\pi'} \geq 0$) in a more or less aggressive way, depending on the strategy.

Unlike \cite{diversity_driven_exploration} who applied proactive and reactive strategies on policies up to 5 timesteps back, we maintain an archive consisting of two clusters seen so far: policies with low and high fitness, respectively. Having this cluster allows to attract/repulse from a set of diverse agents, replacing high-reward policies by policies with similar performance. Indeed, without this process, elements of the archive would collapse on the most frequent policy, from which all agents would attract/repulse. To avoid performing AR against a single "average policy" , we separate low-reward and high-reward agents via clustering.

\section{Related Work}
\label{sec:rel_work}

% \subsection{Density estimation for better exploration}

The challenges of exploration are well studied in the RL literature.
Previously proposed approaches for overcoming hard exploration domains tend to either increase the capacity of the state-action value function~\citep{alpha_div_policy_gradient,gal2016dropout} or the policy expressivity~\citep{touati,boosting_trpo,sac_nf}. This work rather tackles exploration from a diverse multi-agent perspective. Unlike prior population-based approaches for exploration~\citep{cem_rl,erl,novelty_seeking_agents}, which seek diversity through the parameters space, we directly promote diversity in the policy space.
The current work was inspired by \cite{diversity_driven_exploration}, who relied on the KL divergence to attract/repulse from a set of previous policies to discover new solutions. However, in their work, the archive is time-based (they restrict themselves to the $5$ most recent policies), while our archive is built following a diversity-seeking strategy (i.e., niching and policies come from multiple agents).
Notably, \method is different of previously discussed works in that it explores the action space in multiple regions simultaneously, a property enforced through the AR mechanism.

% At a frontier of neuroevolution
The proposed approach bears some resemblance with \cite{stein_variationam_pg}, who took advantage of a multi-agent framework in order to perform repulsion operations among agents using of similarity kernels between parameters of the agents. The AR mechanism gives rise to exploration through structured policy rather than randomized policy. This strategy has also been employed in multi-task learning~\citep{gupta2018meta}, where experience on previous tasks was used to explore on new tasks.

\section{Experiments}
\label{sec:experiments}

% This section first provides the intuition behind the proposed \method using a didactic example before comparing against state-of-the-art methods on control environments.

\subsection{Didactic example}

\begin{figure}
    \centering
    \includegraphics[width=0.5\linewidth]{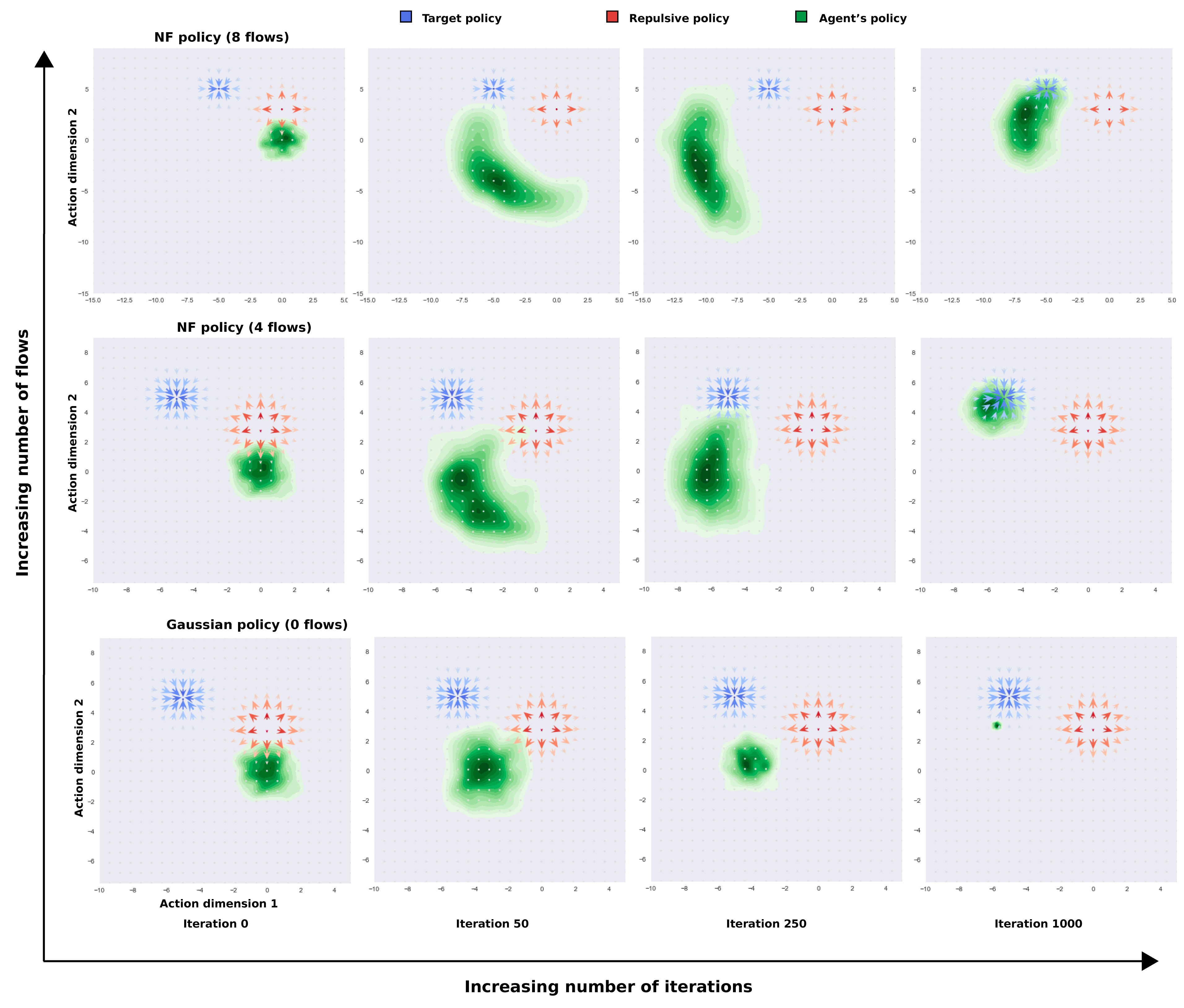}
    \caption{Agent trained to imitate a target while avoiding a repulsive policy using a proactive strategy. Increasing the number of flows leads to more complex policy's shape.}
    \label{fig:mutation_illustration}
\end{figure}

% \thang{shorten this guy and move experiments parameters in appendix, readers just need to understand high lvl intuition, they probably won't look to reproduce this}
Consider a 2-dimensional multi-armed bandit problem where the actions lie in the real square $[-6,6]^2$. We illustrate the example of using a proactive strategy where a SAC agent with radial flows policy imitates a desirable (expert) policy while simultaneously repelling from a less desirable policy. 
% Figure~\ref{fig:mutation_illustration} shows the policy of a SAC agent 
% with $4$ radial flows and randomly initialized weights, 
% starting with actions centered at $(0,0)$.
% \rdh{It would be good to see some different initializations. What if it was centered on the bad mode? Also can we see this without the NF part to show it does worse?}\bogdan{If I have enough time, then I'll run that. Else, I'll have it ready for rebuttal}. 
% All flow parameters are $\ell_1$ regularized with coefficient $2$. 
The task consists in matching the expert's policy (blue density) while avoiding taking actions from a repulsive policy $\pi'$ (red). We illustrate the properties of radial flows in Figure~\ref{fig:mutation_illustration} by increasing the number of flows (where $0$ flow corresponds to a Gaussian distribution).

% shows the evolution of the agent's policy density as a function of training steps. 

We observe that increasing the number of flows (bottom to top) leads to more complex policy's shapes and multimodality unlike the Gaussian policy which has its variance shrinked (indeed, the KL divergence is proportional to the ratio of the two variances, hence maximizing it can lead to a reduction in the variance which can be detrimental for exploration purpose). Details are provided in Appendix.

% combining NF layers with a repulsion term allows the agent to imitate the target policy while avoiding the repulsive policy 
% (a comparison with Gaussian policy is provided in Appendix Section~\ref{sec:didactic_details}).
% \rdh{do we? We don't show this experiment without the NF}.  \thang{yes, if time allows, we ll try to perform this with a gaussian policy} 
% Note however that although the agent becomes identical to the target after 1000 iterations in this example, this is rarely the case in higher-dimensional action-spaces since training a policy until convergence is too costly.

\begin{figure}[t]
    \centering
    \includegraphics[width=0.5\linewidth]{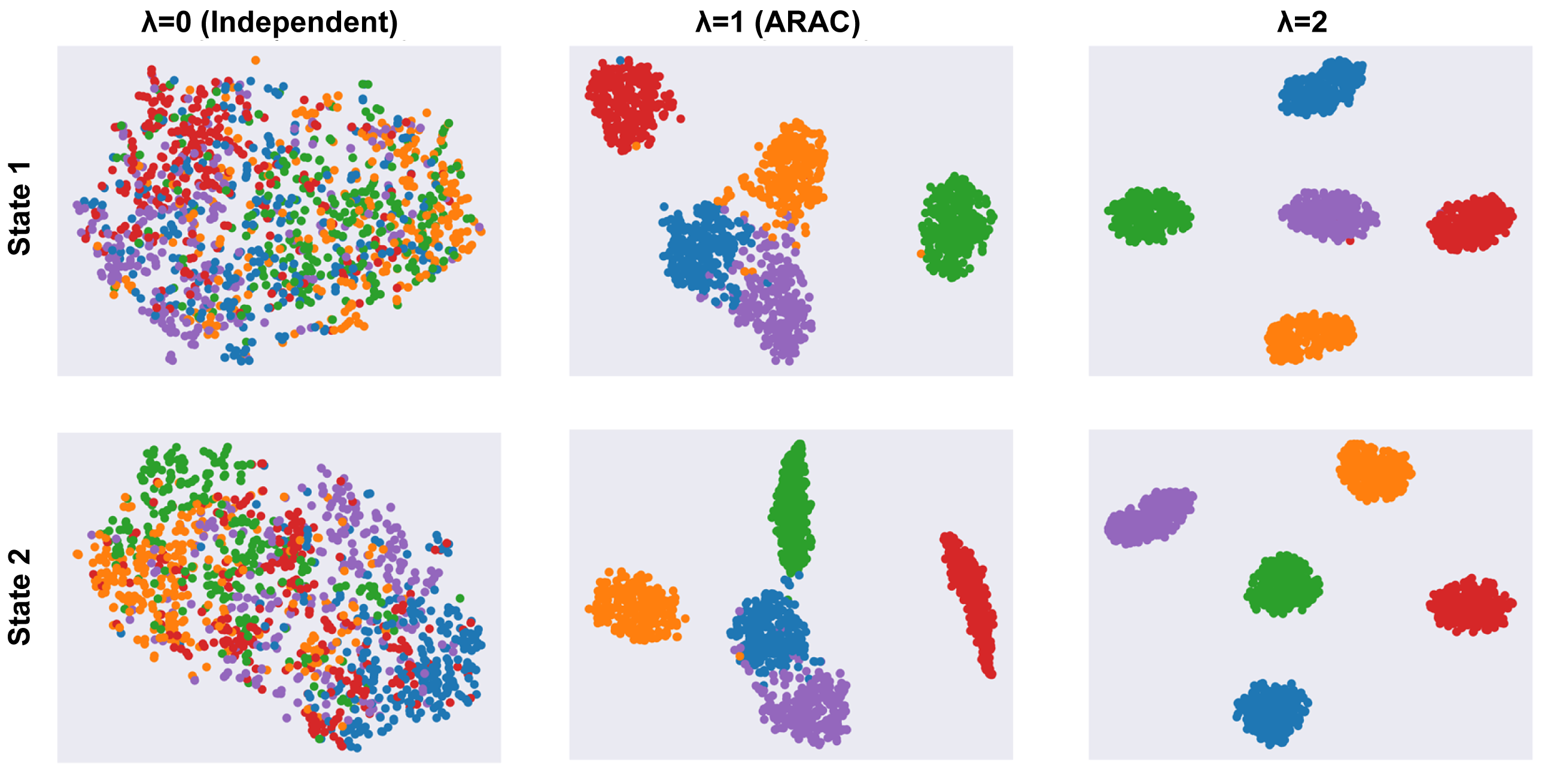}
    \caption{Mapping in two-dimensional space (t-SNE) of agents’ actions for two arbitrary states. Each color represents a different agent.}
    \label{fig:repulsive_forces}
\end{figure}

\subsection{MuJoCo locomotion benchmarks}

We now compare \method against the CEM-TD3~\citep{cem_rl}, ERL~\citep{erl} and CERL \citep{CERL} baselines on seven continuous control tasks from the MuJoco suite~\citep{rllab}: \texttt{Ant-v2}, \texttt{HalfCheetah-v2}, \texttt{Humanoid-v2}, \texttt{HumanoidStandup-v2}, \texttt{Hopper-v2}, \texttt{Walker2d-v2} and \texttt{Humanoid (rllab)}. We also designed a sparse reward environment \texttt{SparseHumanoid-v2}. All algorithms are run over $1$M time steps on each environment, except \texttt{Humanoid (rllab)} which gets $2$M time steps and \texttt{SparseHumanoid-v2} on $0.6$M time steps.

\method performs $R=10$ rollouts for evaluation steps every $10,000$ interaction steps with the environment. We consider a small population of $N=5$ individuals with $K=2$ as elites. Every SAC agent has
% the same architecture as in \citep{sac_nf}, i.e
one feedforward hidden layer of $256$ units acting as state embedding, followed by a radial flow of length $\in \{3,4\}$.
% (We found those number of flows heuristically by adding one more flow than what was previsouly found in \citep{sac_nf}).
A temperature of $\alpha=0.05$ or $0.2$ is used
across all the environments (See appendix for more details).
% , except for \texttt{Humanoid (rllab)} where $\alpha=0.025$. 
AR operations are carried out by sampling uniformly $n=5$ archived policies from $\cG$. Parameters details are provided in the Appendix (Table~\ref{tab:experiments_parameters}).
% \audrey{Should we keep referring to \cite{sac_nf} like this or just give the values here/in Appendix?}
%
All networks are trained with Adam optimizer~\citep{adam} using a learning rate of $3\text{E}^{-4}$. Baselines CEM-TD3\footnote{\url{https://github.com/apourchot/CEM-RL}}, ERL\footnote{\url{https://github.com/ShawK91/erl\_paper\_nips18}}, CERL\footnote{\url{https://github.com/IntelAI/cerl}} use the code contained in their respective repositories.

\begin{table}[ht]
\hspace*{-0.0cm}
    \centering
    {\small \begin{tabular}{c|c|c|c|c|c|c|c}
    \hline
        & \method & CEM-TD3 & CERL & ERL & SAC - NF & SAC & TD3   \\ \hline
     Ant & \textbf{\underline{\SN{6.044e3}}}  & \SN{4.239e3} & \SN{1.639e3}  & \SN{1.442e3} & \SN{4.912e3} & \SN{4.370e3} &  \SN{4.372e3}\\
     HC & \SN{1.0264e4} & \SN{1.0659e4} & \SN{5.703e3}   & \SN{6.746e3}  & \SN{8.429e3}  & \textbf{\underline{\SN{1.19e4}}}  & \SN{9.543e3}  \\ %hc: $7830 \pm 724$, sac-nf:6504 \pm 524
    Hopper & \textbf{\underline{\SN{3.587e3}}} & \textbf{\underline{\SN{3.655e3}}} &  \SN{2.970e3} & \SN{1.149e3}  & \SN{3.538e3}  &  \SN{2.794e3}   &  \SN{3.564e3} \\
     Hu& \textbf{\underline{\SN{5.965e3} }} & \SN{2.12e2} &  \SN{4.756e3}  & \SN{5.51e2}  & \SN{5.506e3}  & \SN{5.504e3}  &  \SN{7.1e1} \\
     Standup& \textbf{\underline{\SN{1.75e5}}} & \SN{2.9e4}  & \SN{1.17e5}  & \SN{1.29e4}  & \SN{1.16e5}    & \SN{1.49e5}  & \SN{5.4e4} \\
      Hu (rllab)  & \textbf{\underline{\SN{1.423e4}}} & \SN{1.334e3} & \SN{3.340e3}  & \SN{5.7e1} & \SN{5.531e3} & \SN{1.963e3}  & \SN{2.86e2} \\
    Walker2d  & \SN{4.704e3}  & \SN{4.710e3} & \SN{4.386e3}  & \SN{1.107e3} & \textbf{\underline{\SN{5.196e3}}} & \SN{3.783e3}  &  \SN{4.682e3}\\
      Hu (Sparse)  &  \textbf{\underline{\SN{8.16e2}}} & \SN{0} &  \SN{1.32e0} & \SN{8.65e0} & \SN{5.47e2}  & \SN{8.8e1}   & \SN{0}\\
         \hline
    \end{tabular}}
    
     \caption{Maximum average return after $1$M ($2$M for \texttt{Humanoid (rllab)} and 600k for \texttt{SparseHumanoid-v2}) time steps 5 random seeds. Bold: best methods when the gap is less than $100$ units. See appendix for average return with standard deviation. Environment short names: HC: \texttt{HalfCheetah-v2}, Hu: \texttt{Humanoid-v2},  Standup: \texttt{HumanoidStandup-v2}}
    \vspace{.5em}
    \label{tab:reported_rewards}
 
\end{table}

Figure~\ref{fig:plot_mujoco} displays the performance of all algorithms on three environments over time steps (see Appendix Figure~\ref{fig:all_mujoco_plot} for all environments). Results are averaged over 5 random seeds. Table~\ref{tab:reported_rewards} reports the best observed reward for each method.

\begin{figure}[h]
    \centering
    \includegraphics[width=0.9\linewidth]{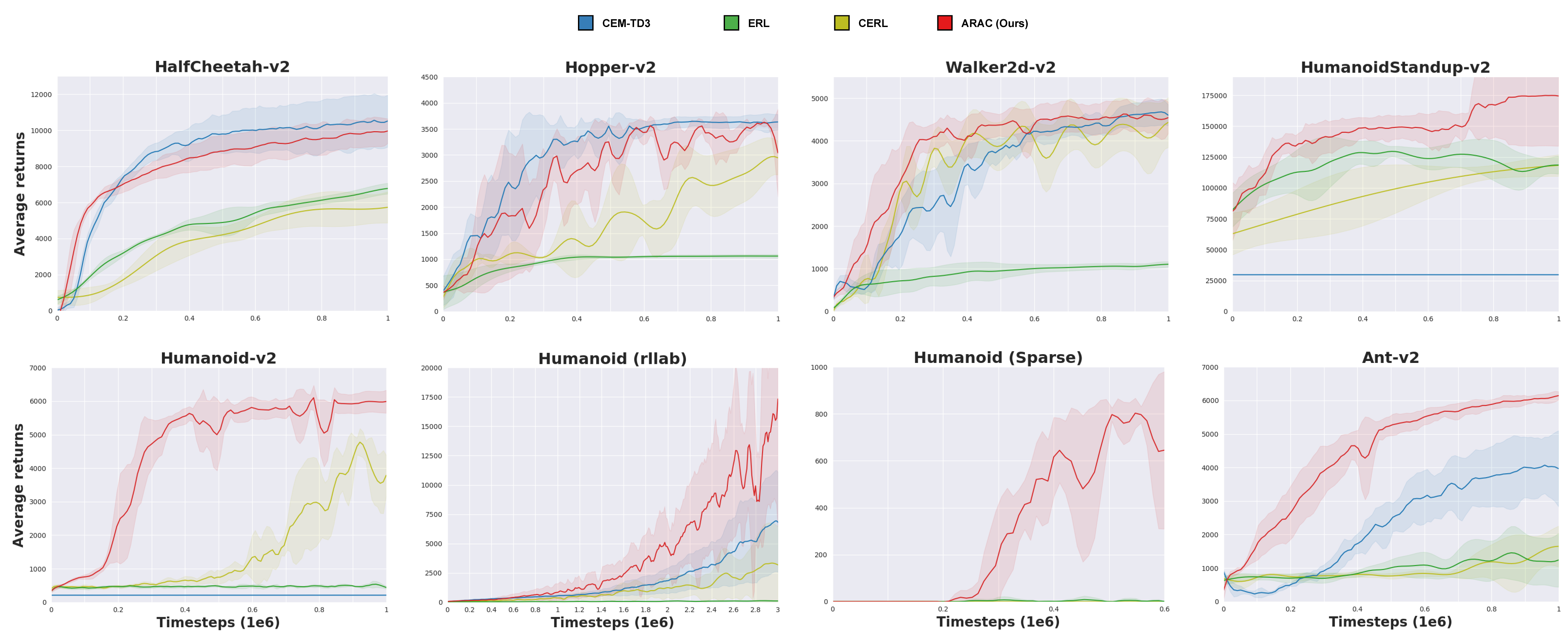}
    \caption{Average return and one standard deviation on 5 random seeds across 8 MuJoCo tasks. Curves are smoothed using Savitzky-Golay filtering with window size of 7.}
    \label{fig:plot_mujoco}
\end{figure}

\paragraph{Small state space environments}

    \texttt{HalfCheetah-v2}, \texttt{Hopper-v2}, and \texttt{Walker2d-v2} are low-dimensional state space environments ($d\leq17$). Except for \texttt{HalfCheetah-v2}, the proposed approach shows comparable results with its concurrent. Those results meet the findings of \citep{parameter_space_noise} that some environments with well-structured dynamics require little exploration. Full learning curves can be found in the Appendix.

\paragraph{Deceptive reward and Large state space environments}

\texttt{Humanoid-v2}, \texttt{HumanoidStandup-v2} and  \texttt{Humanoid (rllab)} belong to bipedal environments with high-dimensional state space ($d=376$ and $d=147$), and are known to trap algorithms into suboptimal solutions. Indeed, in addition to the legs, the agent also needs to control the arms, which may influence the walking way and hence induce deceptive rewards~\citep{novelty_seeking_agents}. Figure~\ref{fig:plot_mujoco} shows the learning curves on MuJoCo tasks. We observe that \method beats both baselines in performance as well as in convergence rate.

\texttt{Ant-v2} is another high-dimensional state space environment ($d\geq 100$). In an unstable setup, a naive algorithm implementing an unbalanced fast walk could still generate high reward, the reward taking into account the distance from start, instead of learning to stand, stabilize, and walk (as expected). 

% In this case, an algorithm can be trapped into a suboptimal solution as falling the farthest

% would give a reward without obviously achieving the goal. 
% Looking at the learning curve, we observe again that \method obtains the best performance, reaching the $4$k milestone after only $300$k steps. This supports the claim that the proposed method allows for efficient exploration strategies by forcing agents to cover complementary state space, as will be shown next.

\paragraph{Sparse reward environment}

To test \method in a sparse reward environment, we created \texttt{SparseHumanoid-v2}. The dynamic is the same as Humanoid-v2 but rewards of +1 is granted only given is the center of mass of the agent is above a threshold (set to 0.6 unit in our case). The challenge not only lies in the sparse reward property but also on the complex body dynamic that can make the agent falling down and terminating the episode. As shown in Figure~\ref{fig:plot_mujoco}, \method is the only method that can achieve non zero performance. A comparison against single agent methods in the Appendix also shows better performance for \method.

\paragraph{Sample efficiency compared with single agent methods} 

Figure~\ref{fig:versus_sac} (in Appendix) also shows that the sample efficiency of the population-based \method compares to a single SAC agent (with and without NFs) and other baselines methods (SAC, TD3). Indeed on \texttt{Humanoid-v2} and \texttt{Ant-v2} \method converges faster, reaching the $6$k ($4$k, respectively) milestone performance after only $1$M steps, while a single SAC agent requires $4$M ($3$M, respectively) steps according to~\citep{sac}. In general, \method achieves competitive results (no flat curves) and makes the most difference (faster convergence and better performance) in the biped environments.

\paragraph{Attraction-repulsion ablation study}
To illustrate the impact of repulsive forces, we introduce a hyperparameter $\lambda$ in the overall loss (Eq.~\ref{eq:sac_pi_ar_loss}):
\begin{align}
    \cL_{\theta,\phi, \lambda}=\cL_{\theta,\phi, \text{SAC}} + \lambda \cL_{\phi,\text{AR}}
\end{align}

We ran an ablation analysis on \texttt{Humanoid-v2} by varying that coefficient. For two random states, we sampled $500$ actions from all agents and mapped these actions onto a two-dimensional space (via t-SNE). Figure~\ref{fig:repulsive_forces} shows that without repulsion ($\lambda=0$), actions from all agents are entangled, while repulsion ($\lambda > 0)$ forces agents to behave differently and hence explore different regions of the action space.

\begin{figure}[h]
    \centering
    \includegraphics[width=0.9\linewidth]{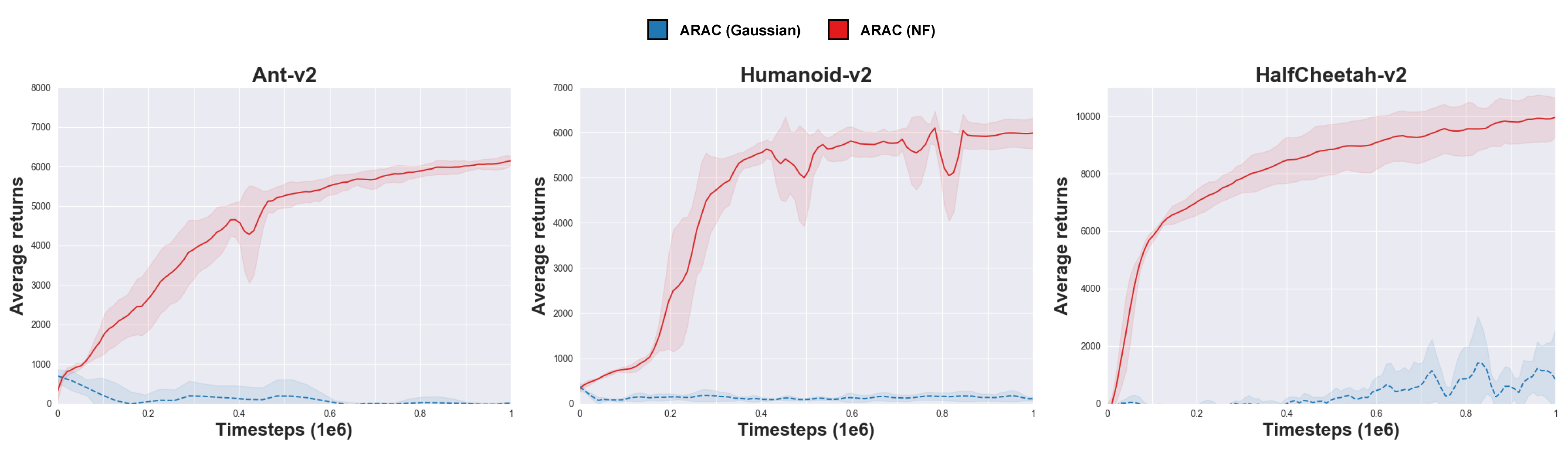}
    \caption{Comparison of \method agents using (1) AR with radial flows, (2) AR with only the base (Gaussian) policy and (3) no AR with radial flows.}
    \label{fig:evo_gaussian}
\end{figure}

The second ablation study is dedicated to highlight the differences between a Gaussian (like in \cite{diversity_driven_exploration} and a NF policy under AR operators. As one can observe in Figure~\ref{fig:evo_gaussian}, using a Gaussian policy deteriorates the solution as the repulsive KL term drives apart the means of agents and blows up/ shrinks the variance of the Gaussian policy. On the other side, applying the AR term on the NF layers maximizes the KL conditioned on the mean and variance of both base policies, resulting in a solution which allows sufficient exploration. More details are provided in the Appendix. 

Also, through a toy example, under a repulsive term, we characterize the policy's shape when increasing the number of our radial flow policy in Figure~\ref{fig:mutation_illustration} (Also show in Appendix). Unlike the diagonal Gaussian policy (SAC) that has symmetry constraint, increasing the number of flows allow radial policy to adopt more complex shape (from bottom to top).
% Same observation can be made for \texttt{Ant-v2}, where a single SAC agent requires $3$M steps to reach the $6$k milestone~\citep{sac} while \method took only $1$M steps.
% single SAC agents need more than $2$M steps (according to \cite{sac}, SAC requires $4$M steps).

%  \begin{figure}
%     \centering
%     \includegraphics[width=0.5\linewidth]{figures/tsne_2frames.png}
%     \caption{$t$-SNE of state visitation after $100$k steps for two methods. \method allows efficient exploration by ensuring a better state space coverage. Each color represents a different agent.}
%     \label{fig:tsne_2frames}
% \end{figure}

% \paragraph{Distributed exploration}

% To illustrate the exploratory benefit of \method, Figure~\ref{fig:tsne_2frames} shows the state visitation after $100$k for \method and CEM-TD3 (mapped into a 2-dimensional space via a $t$-SNE) on the \texttt{Ant-v2} environment, where each color represents a different agent. We observe that \method agents cover the space with minimal overlapping compared with CEM-TD3.
% % The latter samples the agent's weights from the same distribution, thus resulting in similar behavior. On the other hand, \method, through AR, forces agents to explore different regions.
% Full timelines can be found in Appendix (Figure~\ref{fig:tsne_versus_cem}).

\section{Conclusion}
\label{sec:conclusion}

% take home msg: 
% \begin{itemize}
% \item exploration though diversity, 
% \item use AR to boost diversity , 
% \item \thang{talk about NF?}
% \item beat baselines, high sample efficiency
% \item most pronounced on bipede env
% \end{itemize}
In this paper, we introduced a population-based approach for structured exploration leveraging distributional properties of normalizing flows. Our method performs local search by means of Attraction-Repulsion strategies. Performing these operations with NF policies allowed a better handle over local solutions. The AR is done with respect to diverse ancestors across all training steps which are sampled from a bi-modal archive.

% As we evolve a population of multiple agents, we discover new solutions by means of Attraction-Repulsion between the agents of the current population and the policies encountered in previous generations. 

% The minimal overlap of the resulting policies translates in a well distributed and wisely coordinated exploration, allowing the agents to cover complementary regions that are relevant for solving the task. 
Empirical results on the MuJoCo suite demonstrate high performance of the proposed method in most settings. Moreover, in biped environments that are known to trap algorithms into suboptimal solutions, \method enjoys higher sample efficiency and better performance compare to its competitors. 
% Results obtained by SD-AR support the claim that working directly in the action (or policy) space rather than the parameter space yields better empirical performance. 
As future steps, borrowing from multi-objective optimization literature methods could allow one to combine other diversity metrics with the performance objective, to in turn improve the coverage of the solution space among the individuals by working with the corresponding Pareto front~\citep{pareto_niche}.

\section*{Acknowledgements}
We want to thank Compute Canada/Calcul Qu\' ebec and Mila -- Quebec AI Institute for providing computational resources. We also thank Linda Petrini and Lucas Caccia for insightful discussions.

\newpage

 %\bibliography{iclr2020_conference}

\bibliographystyle{iclr2020_conference}

\newpage

\onecolumn
\section*{Appendix}

\subsection*{Reproducibility Checklist}

We follow the reproducibility checklist~\citep{joelle_reproduciblity} and point to relevant sections explaining them here.
\\
For all algorithms presented, check if you include:\\
\begin{itemize}
    \item \textbf{A clear description of the algorithm, see main paper and included codebase.}
    The proposed approach is completely described by Alg.~\ref{alg:evo_nf} (main paper), \ref{alg:rollout} (Appendix), and~\ref{alg:archive} (Appendix).
    % Our algorithm is outlined in the main body of the paper; the rollout and archive update methods are found in the Appendix.
    The proposed population-based method uses attraction-repulsion operators in order to enforce a better policy space coverage by different agents.
    % cover previously unvisited regions of the policy space.
\item \textbf{An analysis of the complexity (time, space, sample size) of the algorithm.}
See Appendix Figure~\ref{fig:all_mujoco_plot} and ~\ref{fig:versus_sac}. Experimentally, we demonstrate improvement in sample complexity as discussed in our main paper. In term of computation time, the proposed method scales linearly with the population size if agents are evaluated sequentially (as presented in Alg.~\ref{alg:evo_nf} for clarity). However, this as mentioned in the paper, can be parallelized. All our results are obtained using $M$ small network architectures with $1\times 256$-units hidden layer followed by $f$ layers of $|A|+2$ units each ($f$ being the number of radial flows and $|A|$ being the action space dimension).
% is slightly higher than the baseline CEM-TD3 because we haven't parallelized the evaluation step.

% Experimentally, we demonstrate
% similarity or improvements on sample complexity as discussed in main paper. In terms of computation time, every agent in our proposed algorithm is lighter than Gaussian agents of the baseline. However, since maintaining a population of agents is more costly than only one policy, the time and space complexity depend on the nature of the environment and the compression rate of the Gaussian agent through NF (cite PREPRINT?).
\item \textbf{A link to a downloadable source code, including all dependencies.} The code is included with the Appendix as a zip file; all dependencies can be installed using Python's package manager. Upon publication, the code would be available on Github.
\end{itemize}
For all figures and tables that present empirical results, check if you include:
\begin{itemize}
    \item \textbf{A complete description of the data collection process, including sample size.} We use standard benchmarks provided in OpenAI Gym (Brockman et al., 2016).
    \item \textbf{A link to downloadable version of the dataset or simulation environment.} See: https://github.com/
\item \textbf{An explanation of how samples were allocated for training / validation / testing.} We do not use a training-validation-test split, but instead report the mean performance (and one standard deviation) of the policy at evaluation time,
openai/gym for OpenAI Gym benchmarks and https://www.roboti.us/index.html for MuJoCo suite.
% averaged across rollouts
obtained with 5 random seeds.
\item \textbf{An explanation of any data that were excluded.} We did not compare on easy environments (e.g. \texttt{Reacher-v2}) because all existing methods perform well on them. In that case, the improvement of our method upon baselines is incremental and not worth mentioning.
% \item \textbf{The range of hyper-parameters considered, method to select the best hyper-parameter configuration, and specification of all hyper-parameters used to generate results.} 
\item \textbf{The exact number of evaluation runs.} 5 seeds for MuJoCo experiments, 1M, 2M or 3M environment steps depending on the domain.
\item \textbf{A description of how experiments were run.} See Section~\ref{sec:experiments} in the main paper and didactic example details in Appendix.
\item \textbf{A clear definition of the specific measure or statistics used to report results.} Undiscounted returns across the whole episode are reported, and in turn averaged across 5 seeds.
\item \textbf{Clearly defined error bars.} Confidence intervals and table values are always mean$\pm$ $1$ standard deviation over 5 seeds.
\item \textbf{ A description of results with central tendency (e.g. mean) and variation (e.g. stddev)}. All results use the mean and standard deviation.
\item \textbf{ A description of the computing infrastructure used.} All runs used 1 CPU for all experiments (toy and MuJoCo) with $8$Gb of memory.
\end{itemize}
\newpage

% \subsection*{Controlling the diversity}
% If $X\sim\mathcal{N}(\mu_X,\Sigma_X)$ and $Y\sim\mathcal{N}(\mu_Y,\Sigma_Y)$ over $\mathbb{R}^d$, then
% \begin{equation}
%     D_{KL}(\mathbb{P}_X||\mathbb{P}_Y)\propto \text{tr}(\Sigma_X^{-1}\Sigma_Y)+(\mu_X-\mu_Y)^\top\Sigma_X^{-1}(\mu_X-\mu_Y)+\log|\Sigma_X|-\log|\Sigma_Y|)
% \end{equation}
% and maximizing $D_{KL}$ with respect to $(\mu_X,\Sigma_X)$ is equivalent to making $\mu_X>>\mu_Y$ for fixed covariance.

% For the radial form of NFs with parameters $\theta = (\alpha,\beta,z)$, taking $X_r=NF_X(\varepsilon)$ and $Y_r=NF_Y(\varepsilon)$ with 1 flow is equivalent to
% \begin{equation}
% \begin{split}
%      \max_{\alpha,\beta,z}D_{KL}(\mathbb{P}_{X_r}||\mathbb{P}_{Y_r})& \propto \max_{\alpha,\beta,z}\mathbb{E}_{\varepsilon}\bigg[(d-1)\log\big(1+\frac{\beta}{\alpha+||\varepsilon-z||})+\log\big(1+\frac{\beta}{\alpha+||\varepsilon-z||}-\beta \frac{\varepsilon-z}{(\alpha+||\varepsilon-z||)^2}\big)\bigg]
%      \label{eq:KL_NF}
% \end{split}
% \end{equation}
% Denote the right hand side function with the expectation term as $g(\alpha,\beta,z)$
% The task is then to show that $\max g <\infty$ 

\subsection*{Impact of repulsive force}
\begin{figure}[h]
    \centering
    \includegraphics[width=\linewidth]{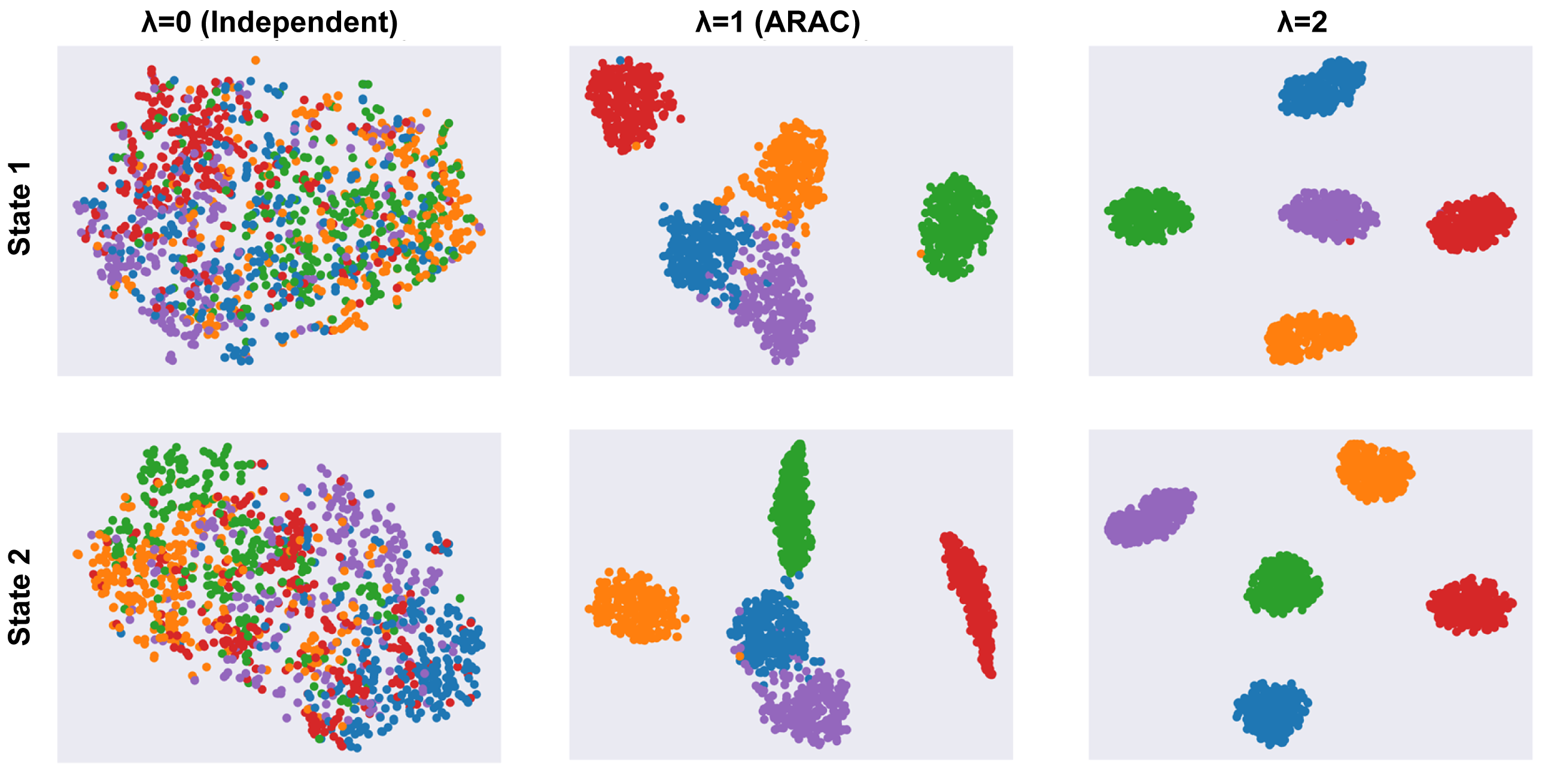}
    \caption{Mapping in two-dimension space (t-SNE) of agents’ actions for two arbitrary states. Each color represents a different agent.}
    \label{fig:my_label}
\end{figure}

To illustrate the impact of the repulsive force coefficient $\lambda$, we ran an ablation analysis by varying that coefficient (recall that the overall loss function is $\cL_{\pi}=\cL_{\pi, \text{SAC}} + \lambda \cL_{\text{AR}}$ where $\lambda=1$ in our experiment). 

For two random states, we sampled $500$ actions from all agents and mapped theses actions in a common 2-dimensional space (t-SNE). 

As shown in the Figure~\ref{fig:my_label}, policies trained without AR ($\lambda=0$) result in entangled actions, while increasing the repulsive coefficient $\lambda$ forces agents to have different actions and hence explore different regions of the policy space. Note that due to the specific nature of t-SNE , the policies are shown as Gaussians in a lower-dimensional embedding, while it is not necessarily the case in the true space.

\newpage

\subsection*{Stabilizing Attraction-Repulsion with Normalizing Flow}

In this section, we illustrate the consequence of the AR operators with a Gaussian policy (as in \cite{diversity_driven_exploration}) and our Normalizing flow policy for \texttt{Ant-v2}, \texttt{Humanoid-v2} and \texttt{HalfCheetah-v2}. As shown in the figure below, AR with Gaussian policies yield worse results. One reason is that the KL term drives apart the mean and variance of the Gaussian policy which deteriorates the main objective of maximizing the reward. On the other side, our method applies the AR only on the NF layers allows enough exploration by deviating sufficiently from the main objective function.

\subsection*{Comparing \method against baselines on Mujoco tasks}

\begin{figure}[h]
    \centering
    \includegraphics[width=\linewidth]{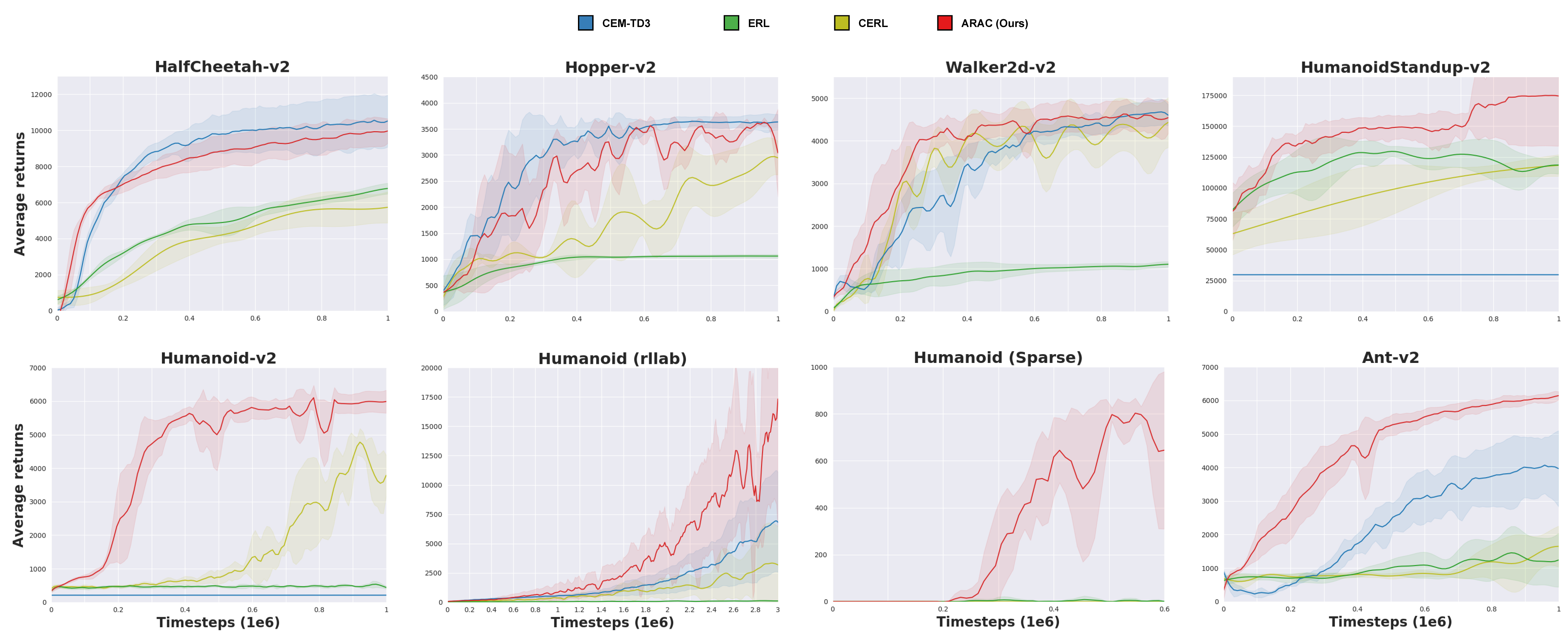}
    \caption{Average return and one standard deviation on 5 random seeds across 7 MuJoCo tasks for \method against baselines. Curves are smoothed using Savitzky-Golay filtering with window size of 7.}
    \label{fig:all_mujoco_plot}
\end{figure}

Figure~\ref{fig:all_mujoco_plot} shows the performance of \method and baselines (CEM-TD3, CERL and ERL) over time steps. Learning curves are averaged over $5$ random seeds and displayed with one standard deviation. Evaluation is done every $10,000$ environment steps using $10$ rollouts per agent. Overall, \method has reasonable performance on all tasks (no flat curves) and demonstrates high performance, especially in humanoid tasks.

\newpage

\subsection*{Benefits of population-based strategies: \method against single agents}

\begin{figure}[h]
    \centering
    \includegraphics[width=\linewidth]{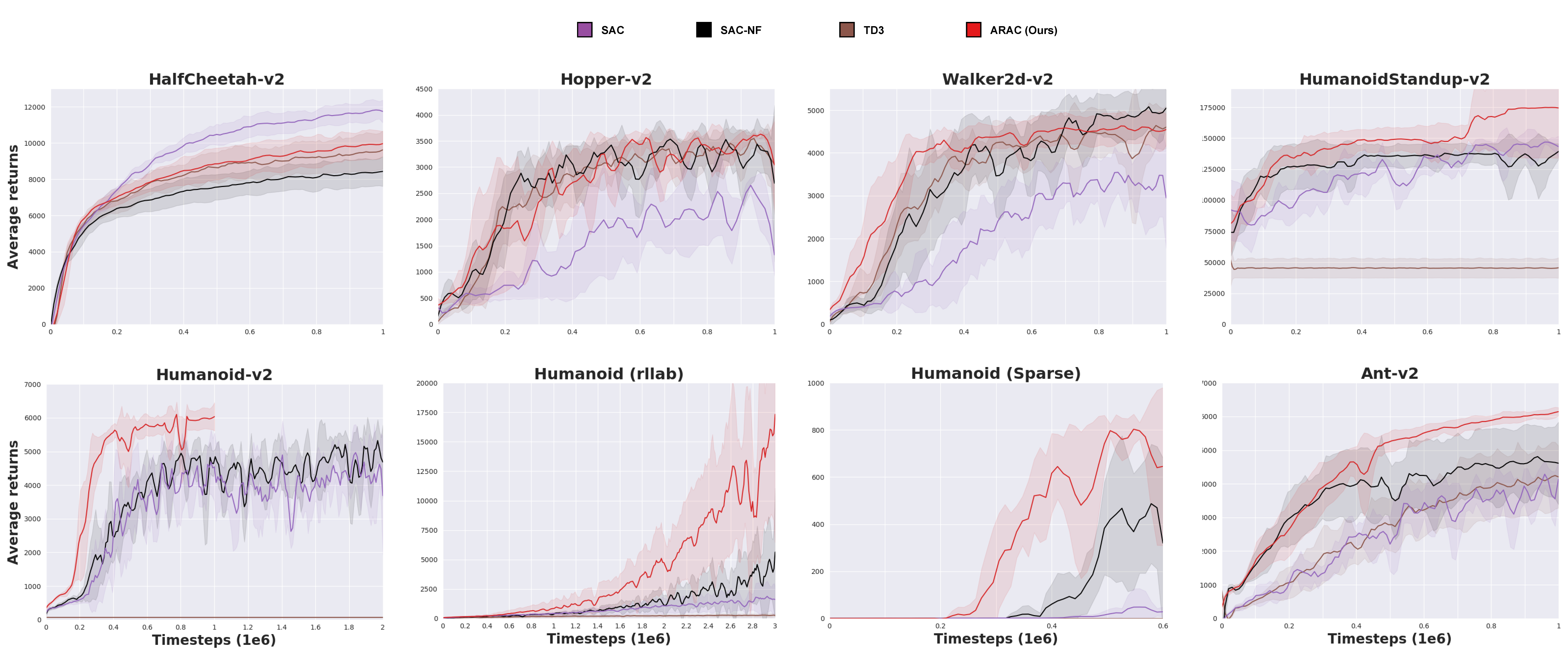}
    \caption{Average return and one standard deviation on 5 random seeds across 7 MuJoCo tasks for \method against single SAC agents (with and without NFs). Curves are smoothed using Savitzky-Golay filtering with window size of 7.}
    \label{fig:versus_sac}
\end{figure}

In this section, we highlight the benefits of the proposed population-based strategy by comparing with single agents. Figure~\ref{fig:versus_sac} shows the performance of \method against a single SAC agent (with and without normalizing flows). Learning curves are averaged over $5$ random seeds and displayed with one standard deviation. Evaluation is done every $10,000$ environment steps using $10$ rollouts per agent. We observe a high beneficial impact on the convergence rate as well as on the performance. \method outperforms single agents in almost all tasks (except for \texttt{HalfCheetah-v2} and \texttt{Walker-v2}) with large improvement. Note the high sample efficiency on humanoid environments ($\texttt{Humanoid-v2}$ and $\texttt{Humanoid (rllab)}$), where it shows faster convergence in addition to better performance. Indeed, on $\texttt{Humanoid (rllab)}$ a single SAC agent reaches the $4$k milestone after $4$M steps~\citep{sac} while \method achieves this performance in less than $2$M steps. Also, in \texttt{SparseHumanoid-v2}, due to its better coordinated exploration, \method could find a good solution faster than SAC-NF.

% \begin{figure}[h]
%     \centering
%     \includegraphics[width=\linewidth]{figures/fig2_mujoco_vs_sac_lightblue-01.png}
%     \caption{In all environments, \method shows a great improvement (in terms of performance as well as in convergence rate) than its counterpart SAC and SAC-NF. Note a high beneficial improvement for humanoid tasks.}
%     \label{fig:versus_sac}
% \end{figure}

% \begin{figure}[h]
%     \centering
%     \includegraphics[width=\linewidth]{figures/fig2_mujoco_vs_sac_olive-01.png}
%     \caption{In all environments, \method shows a great improvement (in terms of performance as well as in convergence rate) than its counterpart SAC and SAC-NF. Note a high beneficial improvement for humanoid tasks.}
%     \label{fig:versus_sac}
% \end{figure}

% \begin{figure}[h]
%     \centering
%     \includegraphics[width=\linewidth]{figures/fig2_mujoco_vs_sac-01.png}
%     \caption{In all environments, \method shows a great improvement (in term of performance as well as in convergence rate) than its counterpart SAC and SAC-NF. Note a high beneficial improvement for humanoid tasks.}
%     \label{fig:versus_sac}
% \end{figure}

\newpage

\subsection*{Overall performances on Mujoco tasks}

\begin{table}[h!]
\hspace*{-2.8cm}
    \centering
    {\small \begin{tabular}{c|c|c|c|c|c|c|c}
    \hline
        & \method & CEM-TD3 & CERL & ERL & SAC - NF & SAC & TD3   \\ \hline
     Ant-v2 & \textbf{6,044 $\pm$ 216} & $4,239 \pm 1,048$ & $1,639 \pm 564$  & $1,442 \pm 819$ & $4,912 \pm 954$  & $4,370\pm 173$ &  $4,372 \pm 900$\\
     HalfCheetah-v2& $10,264 \pm 271$ & 10,659 $\pm$ 1,473 & $5,703 \pm 831$  & $6,746 \pm 295$ & $8,429 \pm 818$  & \textbf{11,896 $\pm$ 574}  & $9,543 \pm 978$ \\ %hc: $7830 \pm 724$, sac-nf:6504 \pm 524
    Hopper-v2 & \textbf{3,587 $\pm$ 65} & \textbf{3,655 $\pm$ 82} &  $2,970 \pm 341$ & $1,149 \pm 3$ & $3,538 \pm 108$ &  $2,794 \pm 729$  &  $3,564 \pm 114$\\
     Humanoid-v2& \textbf{5,965 $\pm$ 51} & $212 \pm 1$ &  $4,756 \pm 454$ & $551 \pm 60$ & $5,506 \pm 147$ & $5,504 \pm 116$ &  $71 \pm 10$ \\
     HumanoidStandup-v2& \textbf{175k $\pm$ 38k} & $29 k \pm 4k$ & $117k \pm 8k$  & $129 k \pm 4k$ & $116 k \pm 9k$  & $149 k \pm 7k$ & $54k \pm 24k$ \\
      Humanoid (rllab)  & \textbf{14,234 $\pm$ 7251} & $1,334 \pm 551$ & $3,340 \pm 3,340$ & $57 \pm 17$ & $5,531 \pm 4,435$ & $1,963 \pm 1,384$ & $286\pm 151$ \\
    Walker2d-v2  & 4,704 $\pm$ 261 & 4,710 $\pm$ 320 & 4,3860 $\pm$ 615 & $1,107 \pm 60$ & \textbf{5,196 $\pm$ 527} & 3,783 $\pm$ 366  &  4,682 $\pm$ 539\\
      SparseHumanoid-v2  &  \textbf{816 $\pm$ 20} & $0\pm 0$ &  $1.32 \pm 2.64
      $ & $8.65 \pm 15.90$ & 547 $\pm$ 268 & 88 $\pm$ 159  & $0 \pm 0$\\
         \hline
    \end{tabular}}
    
     \caption{Maximum average return after $1$M ($2$M for \texttt{Humanoid (rllab)} and 600k for \texttt{SparseHumanoid-v2}) time steps $\pm$ one standard deviation on 5 random seeds. Bold: best methods when the gap is less than $100$ units.}
    \vspace{.5em}
    % \label{tab:reported_rewards}
 
\end{table}

\begin{table}[h]
\hspace*{-2.0cm}
    \centering
    {\small \begin{tabular}{c|c|c|c|c|c|c|c}
    \hline
        & CLEAR &  TRPO  & PPO &Trust-PCL & \cite{plappert2017parameter} & \cite{touati} & \cite{diversity_driven_exploration}\\ \hline
     HalfCheetah-v2 & $\mathbf{10,264}$ &   $-15$  & $2,600$ &   $2,200$ & $5,000$ & $7,700$ & $4,200$ \\
      Walker-v2 & $\mathbf{4,764}$ & $2,400$  & $4,050$ &   $400$ & $850$ & $500$ & N/A  \\
      Hopper-v2 & $\mathbf{3,588}$ & $600$    & $3,150 $ & $280$ &  $2,500$ & $400$ &  N/A \\
     Ant-v2 & $\mathbf{6,044}$ &  $-76$   &  $1,000$ &    $1,500$ & N/A & N/A & N/A \\
     Humanoid-v2 & $\mathbf{5,939}$ &  $400$  & $400$  &     N/A   & N/A  & N/A & $1,250$ \\
     HumanoidStandup-v2& $\mathbf{163,884}$ & $80,000$ &  N/A &   N/A   & N/A  & N/A & N/A \\
      Humanoid (rllab)  & $\mathbf{4,117}$  & $23$ & $200$   &  N/A   & N/A & N/A & N/A \\
         \hline
    \end{tabular}}
    % \vspace{.5em}
    \caption{Performance after 1M (except for rllab which is 2M) timesteps on $5$ seeds. Values taken from their corresponding papers. N/A means the values were not available in the original paper.
     }
    \label{tab:reported_rewards_touati}
\end{table}

\newpage 

\subsection*{Experimental parameters}

Table~\ref{tab:experiments_parameters} provides the hyperparameters of \method used to obtain results in the MuJoCo domains. The noise input for normalizing flows in SAC policies (see Sec.~\ref{sec:prelim:nfs}) is sampled from $\mathcal{N}(0,\sigma)$, where the variance $\sigma$ is a function of the state (either fixed at a given value or learned). 

% Note that for non-humanoid tasks, we found out that fixing $\sigma$ tends to yield more stable agent behavior due to the stochasticity induced by AR operators in the population-based setting. Humanoid tasks have a high-dimensional state space ($d\geq 375$) and hence there might be benefit in learning the noise variance as a function of each specific state in order to control exploration.
% Note that for the flow value, we added one more flow compared to the value in \cite{sac_nf}.

\begin{table}[H]
\centering
\resizebox{0.6\textwidth}{!}{
\begin{tabular}{l|c|c|c|c|c|c}
 \cline{1-7}
    \multicolumn{7}{c}{\method parameters}         \\  \hline
                      & $\#$ flows  &  $\sigma$ & $G$ & $p$ & alpha  &  strategy \\\cline{1-7}
\multicolumn{1}{c|}{ Ant-v2}    & $3$   &  $0.2$  & $10$ &  $1$ &  0.2  & proactive  \\ 
\multicolumn{1}{c|}{ HalfCheetah-v2}           & $4$   & $0.4$  & $20$ &  $2$ & 0.2 & proactive  \\ 
\multicolumn{1}{c|}{ Hopper-v2 }     &  $4$ & $0.8$ & $20$ &  $1$  & 0.05  & proactive  \\  
\multicolumn{1}{c|}{ Walker2d-v2}     & $4$ &    $0.6$ & $10$ &  $3$  & 0.05 &  proactive \\  
\multicolumn{1}{c|}{ Humanoid-v2}     & $3$ &  $0.6$  & $10$ &  $1$  & 0.05  & reactive \\   
\multicolumn{1}{c|}{ HumanoidStandup-v2}     & $3$ &  $\sigma$  & $20$ &  $1$  & 0.2 &  reactive \\   
\multicolumn{1}{c|}{ Humanoid (rllab)}     & $3$ &    $\sigma$ & $10$ & $1$    & 0.05  & proactive \\  
\multicolumn{1}{c|}{ SparseHumanoid-v2}     & $2$ &    $0.6$ & $20$ & $1$    & 0.2  & proactive \\ 
\hline
  \multicolumn{5}{c}{Adam Optimizer parameters} \\  \hline
\multicolumn{1}{c|}{ $\alpha_{\gamma}$}     & \multicolumn{1}{|c}{$3.10^{-4}$}  \\
\multicolumn{1}{c|}{ $\alpha_{\omega}$}     & \multicolumn{1}{|c}{$3.10^{-4}$}  \\
\multicolumn{1}{c|}{ $\alpha_{\theta}$}     & \multicolumn{1}{|c}{$3.10^{-4}$}  \\
\multicolumn{1}{c|}{ $\alpha_{\phi}$}     & \multicolumn{1}{|c}{$3.10^{-4}$}  \\ \hline
 \multicolumn{5}{c}{Algorithm parameters}         \\  \hline
 \multicolumn{1}{c|}{ Batch size $m$}     & \multicolumn{1}{|c}{$256$}  \\
 \multicolumn{1}{c|}{ Buffer size $\mathcal{B}$}     & \multicolumn{1}{|c}{$10^6$}  \\
 \multicolumn{1}{c|}{ Archive sample size $n$}     & \multicolumn{1}{|c}{$5$} \\\hline
\end{tabular}}
\caption{\method parameters.}
\label{tab:experiments_parameters}
\end{table}

\newpage
\subsection*{Impact of number of flows on the policy shape}
\label{sec:didactic_details}

\begin{figure}[h]
    \centering
    \includegraphics[width=\linewidth]{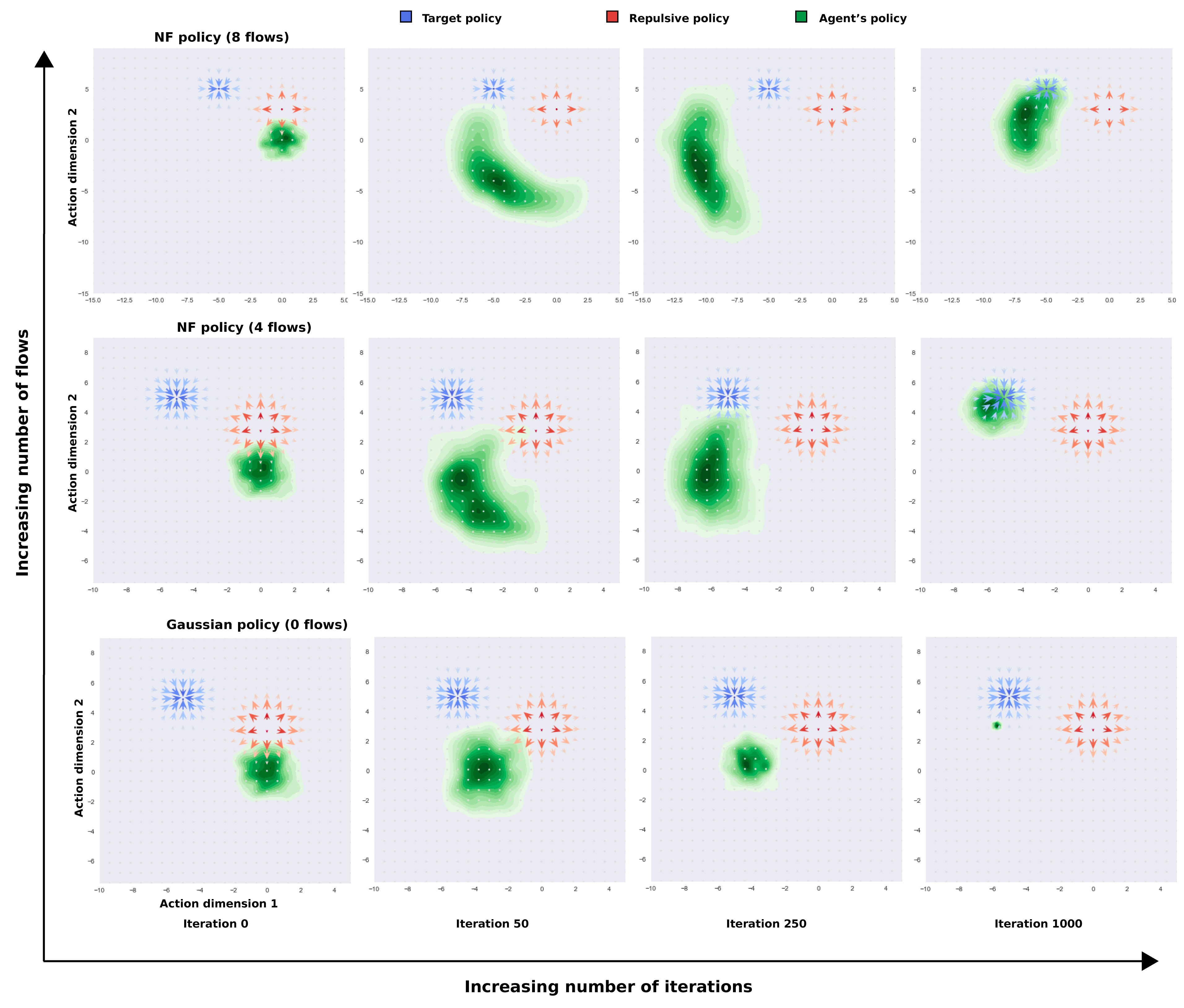}
    \caption{Single state didactic illustration of attraction-repulsion operators. Comparing behavior of NF policy against Gaussian policy with learned variance under a repulsive constraint.}
    \label{fig:full_toy}
\end{figure}

% The didactic toy setting is a 2-dimensional multi-armed bandit problem where the actions lie in the real square $[-6,6]^2$. The task consists in matching a desirable (expert) policy while simultaneously repelling from a less desirable policy.
We used a single SAC agent with different radial flows numbers and randomly initialized weights, starting with actions centered at $(0,0)$. All flow parameters are $\ell_1$ regularized with hyperparameter $2$. The agent is trained with the classical evidence lower bound (ELBO) objective augmented with the AR loss (Eq.~\ref{eq:attraction_repulsion}), where the coefficient of the repulsive policy $\pi'$ is given by $\beta_t=\frac{10}{t+1}$.
Fig.~\ref{fig:full_toy} shows how both the NF and learned variance Gaussian policies manage to recover the target policy.
We see that NF takes advantage of its flexible parametrization to adjust its density and can show asymmetric properties unlike the Gaussian distribution. This indeed can have advantage in some non symmetric environment where the Gaussian policy would be trapped into a suboptimal behavior.
Finally, increasing the number of flows (from bottom to top) can lead to more complex policy's shape.
% \thang{understandable?}
% We see that the former takes advantage of its flexible parameterization to adjust the density to avoid repulsive regions while still following the target; this leads to a faster convergence rate since not only the location, but also the shape of the policy is allowed to change. 
% Moreover, in this precise example, the flow-based policy avoids mode collapse unlike the Gaussian policy, which reduces to a Dirac distribution on the mode of the target policy.

\newpage
\subsection{Pseudo-code for \method}
\begin{algorithm}
   \caption{\method: Attraction-Repulsion Actor-Critic}
   \label{alg:evo_nf}
\begin{algorithmic}[1]
   \STATE {\bfseries Input:} population size $M$; number of elites $K$; maximum archive capacity $G$; archive sample size $n$;  number of evaluation rollouts $R$; actor coefficient $p$; strategy (either proactive or reactive).
   \newline
   
   \STATE Initialize value function network $V_\nu$ and critic network $Q_\omega$
   \STATE Initialize population of policy networks $\{ \pi_{\phi,\theta}^{m} \}_{m=1}^{M}$
   \STATE Initialize empty archive $\cG$ and randomly assign $K$ individuals to top-$K$ \tikzmark{right}
   \newline

   \STATE $\text{total\_step} \leftarrow 0$
   \WHILE{$\text{total\_step} \leq \text{max\_step}$}
   \STATE $\text{step} \leftarrow 0$
   
   \tikzmark{top_samples}
   \FOR{agent $m=1\dots M$}\label{alg:ln:beg_samples}
   \STATE $( \_, \text{step}\, s)\leftarrow \rollout(\pi^{m},\withnoise,\episode{1})$
   \STATE $\text{step} \leftarrow \text{step} + s$
   \STATE $\text{total\_step} \leftarrow \text{total\_step} + s$
   \ENDFOR\label{alg:ln:end_samples}
   \tikzmark{bottom_samples}
   
   \tikzmark{top_critic}
   \STATE $C=\text{step}/K$\label{alg:ln:beg_update_critic}
   \FOR{policy $\pi^e$ in top-$K$} 
   \STATE Update critic with $\pi^{e}$ for $C$ mini-batches (Eq.~\ref{eq:critic_loss})
   \STATE Update value function (Eq.~\ref{eq:v_loss})
   \ENDFOR\label{alg:ln:end_update_critic}
   \tikzmark{bottom_critic}
   
   \tikzmark{top_actors}
   \FOR{agent $m=1\dots M$}\label{alg:ln:beg_update_actors}
   \IF {policy $\pi^m$ is in top-$K$} 
   \STATE Sample $n$ archived policies uniformly from $\cG$\label{alg:ln:beg_ar}
%   \rdh{ancestor of who?} \thang{our archive contains K copies of past policies and we sample randomly 5 of them} \audrey{Changed ``ancestors'' for ``policies''.}
   \STATE Update actor $\pi^{m}$ for $\frac{\text{step}}{M}. p$ mini-batches (Eq.~\ref{eq:sac_pi_ar_loss} and~\ref{eq:proactive} or~\ref{eq:reactive})\label{alg:ln:end_ar}
   \ELSE
   \STATE Update actor $\pi^{m}$ for $\frac{\text{step}}{M}. p$ mini-batches (Eq.~\ref{eq:sac_pi_loss})
   \ENDIF
   \ENDFOR\label{alg:ln:end_update_actors}
   \tikzmark{bottom_actors}
   
   \tikzmark{top_eval}
   \FOR{agent $m=1\dots M$}\label{alg:ln:beg_eval}
   \STATE $(\text{Fitness}_m ,\_)\leftarrow \rollout(\pi^m,\withoutnoise,\episodes{R})$
   \ENDFOR
   \tikzmark{bottom_eval}\label{alg:ln:end_eval}

   \STATE Rank population $\{ \pi_{\phi,\theta}^{m} \}_{m=1}^{M}$ and identify top-$K$\label{alg:ln:topk}
   \STATE $\updatearchive(\cG, \{\pi^m_{\phi,\theta}\}_{m=1}^M, G)$\label{alg:ln:archive}
    \ENDWHILE
\end{algorithmic}
\AddNote{top_samples}{bottom_samples}{right}{\emph{Collect samples}}
\AddNote{top_critic}{bottom_critic}{right}{\emph{Update critic}}
\AddNote{top_actors}{bottom_actors}{right}{\emph{Update actors}}
\AddNote{top_eval}{bottom_eval}{right}{\emph{Evaluate actors}}
\end{algorithm}

\newpage
\subsection*{Complementary pseudo-code for \method}

Algorithms~\ref{alg:rollout} and~\ref{alg:archive} respectively provide the pseudo-code of functions \texttt{rollout} and \texttt{update$\_$archive} used in Algorithm~\ref{alg:evo_nf}.

\begin{algorithm}[H]
   \caption{\texttt{rollout}}
   \label{alg:rollout}
\begin{algorithmic}
   \STATE {\bfseries Input:} actor $\pi$; noise status; number of episodes $E$; replay buffer $\mathcal{B}$;
%   \STATE $\text{it} \leftarrow 0 $
   \STATE $\text{Fitness} \leftarrow 0 $
   \FOR{$\text{episode}=1,\dots,E$} 
   \STATE $\vec{s}  \gets \text{Initial state }\vec{s}_0\text{ from the environment}$ 
    \FOR{step $t=0\dots$ termination}
    \IF {with noise}
    \STATE Sample noise $z$
    \ELSE 
    \STATE Set $z \leftarrow 0$
    \ENDIF
   \STATE $\vec{a}_t \sim \pi(.|\vec{s}_t,z)$ 
    % \STATE $\text{step} \leftarrow \text{step}+1 $
   \STATE Observe $\vec{s}_{t+1} \sim P(\cdot|\vec{s}_t,\vec{a}_t)$ and obtain reward $r_t$
   \STATE $\text{Fitness} \leftarrow  \text{Fitness} + r_t $
   \STATE Store transition ($\vec{s}_t$,$\vec{a}_t$,$r_t$,$\vec{s}_{t+1}$) in $\mathcal{B}$ 
   \ENDFOR
   \ENDFOR
   \STATE $\text{Fitness} \leftarrow  \text{Fitness} / E$
   \RETURN Average fitness per episode and number of steps performed
 
\end{algorithmic}
\end{algorithm}

\begin{algorithm}[H]
   \caption{\texttt{update$\_$archive}}
    \label{alg:archive}
\begin{algorithmic}
   \STATE {\bfseries Input:} archive $\cG$; population of size $M$; maximal archive capacity $G$.
   \IF{$|\cG|< G$}
        \STATE Add all agents of current population to $\cG$
   \ELSE
       \STATE $c_1,c_2\leftarrow 2\text{-means}(\text{fitness of individuals in~} \cG)$
       \FOR{agent $m=1,\dots,M$}
       \STATE Assign agent $m$ to closest cluster $c\in \{c_1,c_2\}$ based on its fitness
       \STATE Sample an archived agent $j\sim \text{Uniform}(c)$
       \STATE Replace archived individual $j$ by $m$
   \ENDFOR
   \ENDIF
   \RETURN Updated archive $\cG$

\end{algorithmic}
\end{algorithm}

\end{document}